\documentclass[11pt,a4paper]{article}
\usepackage{geometry}
\usepackage{coling2020}
\usepackage{times}
\usepackage{underscore}
\usepackage{url}
\usepackage{latexsym}
\usepackage{multicol}
\usepackage{graphicx}
\usepackage{graphics}
\usepackage{microtype}
\usepackage{bm}
\usepackage{url}
\usepackage{hyperref}
\usepackage{caption}
\usepackage{booktabs}
\colingfinalcopy 
\title{Will\_Go at SemEval-2020 Task 3:\\An Accurate Model for Predicting the (Graded) Effect of Context in Word Similarity based on BERT}

\author{Wei Bao\footnotemark[1]\\
  Southeast University \\
  {\tt willinseu@gmail.com} \\\And
  Hongshu Che\footnotemark[1] \\
  Southeast University \\
  {\tt hsche1222@gmail.com} \\\And
  Jiandong Zhang\footnotemark[1] \\
  Southeast University \\
  {\tt zjdx1998@gmail.com} \\
  }

\begin{document}
\maketitle

\begin{abstract}
  Natural Language Processing (NLP) has been widely used in the semantic analysis in recent years. Our paper mainly discusses a methodology to analyze the effect that context has on human perception of similar words, which is the third task of SemEval 2020. We apply several methods in calculating the distance between two embedding vector generated by Bidirectional Encoder Representation from Transformer (BERT). Our team \textit{will\_go} won the 1st place in Finnish language track of subtask1, the second place in English track of subtask1. 
\end{abstract}
\section{Introduction}
\label{intro}

%
%
\blfootnote{
    %
    %
    \hspace{-0.65cm}  
    %
    %
    %
    %
    \hspace{-0.65cm}  
    This work is licensed under a Creative Commons 
    Attribution 4.0 International License.
    License details:
    \url{http://creativecommons.org/licenses/by/4.0/}.
}

\renewcommand{\thefootnote}{\fnsymbol{footnote}}
\footnotetext[1]{These authors contributed equally to this work.}
\renewcommand{\thefootnote}{\arabic{footnote}}

Computing the meaning difference between words in the semantic level is a task that has been widely discussed. In the area of natural language processing (NLP) like information retrieval (IR), there are many specific applications using similarity, such as text memorization \cite{lin-summary}, text categorization \cite{categorization}, Text Q\&A \cite{QA}, etc.

The task3 of SemEval-2020\footnote{https://competitions.codalab.org/competitions/20905} focuses on the influence of context when humans perceive similar words. As we all know, polysemous words have different meanings in a totally different context, which the current translation system can recognize very well. However, many translation systems can't exactly predict the subtle variance on the meanings of words, which is also caused by a different context. 

Task3 has two sub-tasks. In subtask1, we are required to predict the extent of change in scores of similarity between two words in different contexts by human annotators. In subtask2, we only predict the absolute score as is in the subtask1 rather than the difference in scores, and we would only discuss subtask1.

Our team uses different algorithms to calculate the distance between two embedding vectors generated by BERT \cite{BERT} and defines it as  the similarity. So we can get the change in similarity by subtraction between two distances. However, this methodology did not get exciting performance in the task evaluation, so we improve this by blending different BERT models, which we will introduce later in Section \ref{overview}. 

\section{Related Work}
There are many methods and models to estimate the similarity between long paragraphs. Most of them treat it as a binary classification problem,  Hatzivassiloglou et al. \cite{longtext99} compute the linguistic vector of features including primitive features and composite features, then they build criteria by feature vectors to classify paragraphs. As for similarity between short sentences, Foltz et al.\cite{foltz1998measurement} suggest a method that provides a characterization of the extent of semantic similarity between two adjacent short sentences by comparing their high-dimensional semantic vectors, which is also a Latent Semantic Analysis (LSA) model. Both LSA and Hyperspace Analogues to Language (HAL) \cite{HALburgess1998explorations} are all corpus-based model, the latter one uses lexical co-occurrence to generate high-dimensional semantic vectors set, where words in this set can be represented as a vector or high-dimensional point so that their similarities can be measured by computing their distances.

Although computing similarity between words are less difficult than between texts, there still exist some sophisticated problems. Similarity between words is not only in morphology but more significantly in semantic meaning. The first step of reckoning the similarity between words is using Word2Vec\cite{word2vecmikolov2013efficient}, which is a group of corpus-based models to generate word embedding, and mainly utilizes two architectures: continuous bag-of-words (CBOW) and continuous Skip-gram. In the CBOW model, the distributed representations of context are made as input to the model and predict the center words, while the Skip-gram model uses the center words as its input and predict the context, which predicts one word in many times to produce several context words. Therefore, the Skip-gram model can learn efficiently from context and performs better than the CBOW model, but it takes much more consumption in training time than the CBOW model. But since hierarchical softmax and negative sampling \cite{mikolov2013distributed} were proposed to optimize the Skip-gram model when training large-scale data.

Word2Vec cannot be used for computing similarity between polysemous words because it generates only one vector for a word, while Embedding from Language Model (ELMo) \cite{elmopeters2018deep} inspired by semi-supervised sequence tagging \cite{peters2017semisupervised} can handle this issue. ELMo is consist of bidirectional LSTM \cite{lstm}, which makes the ELMo have an understanding of both next and previous word, it obtains  contextualized word embedding by weight summation over the output of hidden layers. Compared with the LSTM used in ELMo, Bidirectional Encoder Representation from Transformer (BERT) \cite{BERT} is a stack of Transformer Encoder \cite{transformervaswani2017attention}, which can be computed in parallel ways and save much time in training. There are two BERT versions with different size, one is BERT Base, which has 12 encode layers with 768 hidden units and 12 attention heads, and the other is BERT Large, which has 24 encode layers with 1024 hidden units and 16 attention heads, achieved state-of-the-art results according to that paper. 

\section{System Overview}
\subsection{Data}
\label{dataset}
The source of our test data is from the CoSimLex dataset \cite{datasetarmendariz2019cosimlex}, which is based on the well known SimLex999 dataset \cite{hill2014simlex999} and provides pairs of words. 

In task3, the English dataset consists of 340 pairs; the Finnish, Croatian, Slovenian consist of 24, 112, 111 pairs respectively.  Here is the quantity count table.

\begin{table}[h]
    \centering
    \begin{tabular}{ccc}
         \toprule
         Language& Abbr. & Count \\
         \midrule
         English & En & 340 \\
         Croatian & HR &  112 \\
         Finnish & FI & 24 \\
         Slovenian & SL & 111 \\
         \bottomrule
    \end{tabular}
    \captionof{table}{Test Dataset in Task3 from CoSimLex dataset}
    \label{tab:dataset}
\end{table}
Each language datafile has eight columns, namely \textit{word1, word2, context1, context2, word1_context1, word2_context1, word1_context2, word2_context2}, and their meanings are first word, second word, first context, second context, the first word in the first context, the second word in the first context, the first word in the second context, the second word in the second context respectively. In addition, word1 and word2 may have a lexical difference between word1_context and word2_context.
\subsection{Methodology}
\label{overview}
The BERT model architecture is based on a multilayer bidirectional Transformer as Figure 1.Instead of the traditional left-to-right language modeling objective, BERT is trained on two tasks: predicting randomly masked tokens and predicting whether two sentences follow each other. BERT model gets a lot of state-of-the-art performance in many tasks, and we also use the BERT model in our strategy. We approach this task as one of tasks that calculates the similarity between two words. In our model, context data would be firstly added into BERT like the following Figure 2.
\begin{figure}[htbp]
\centering
    \begin{minipage}[t]{0.48\textwidth}
        \centering
        \includegraphics[width=0.90\linewidth]{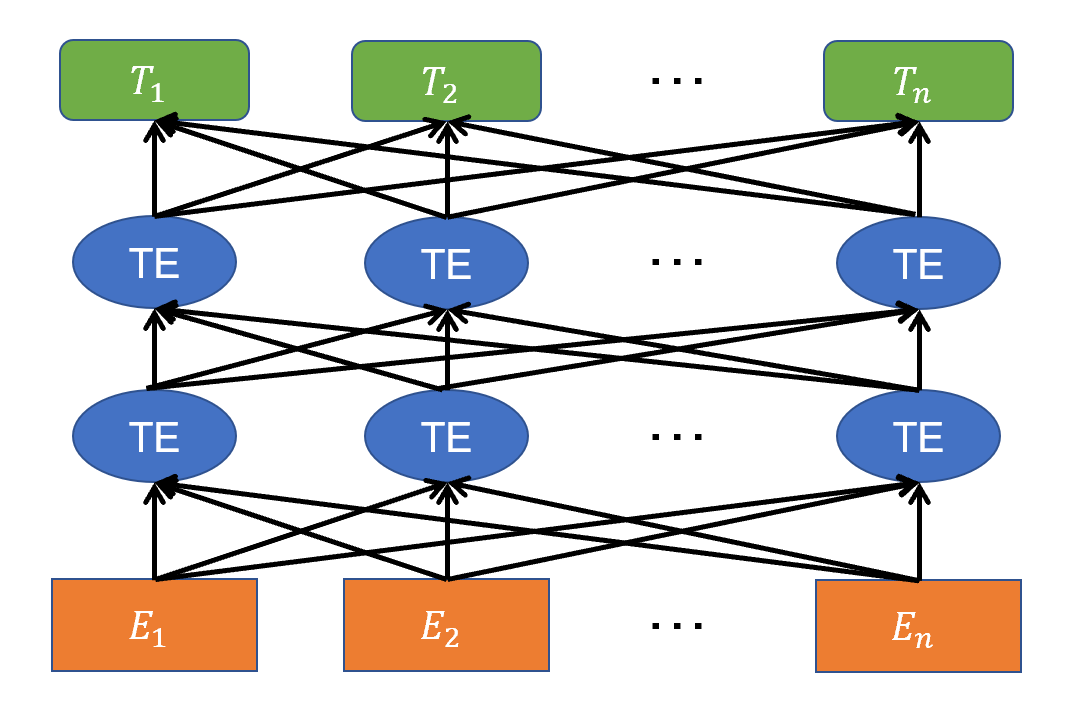}
        \caption{Bidirectional Transformer architectures of BERT}
        \end{minipage}
    \begin{minipage}[t]{0.48\textwidth}
        \centering
        \includegraphics[width=0.60\linewidth]{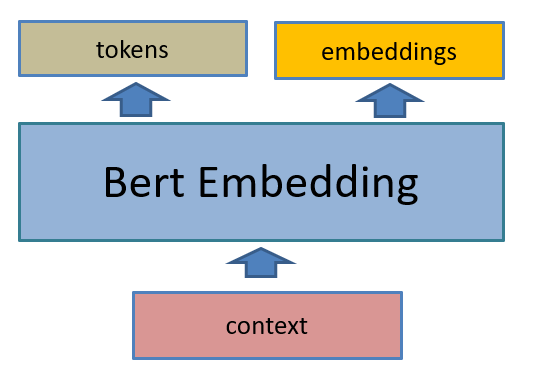}
        \caption{BERT schematic diagram}
    \end{minipage}
\end{figure}



\begin{figure}
    \centering
    \includegraphics[width=\linewidth]{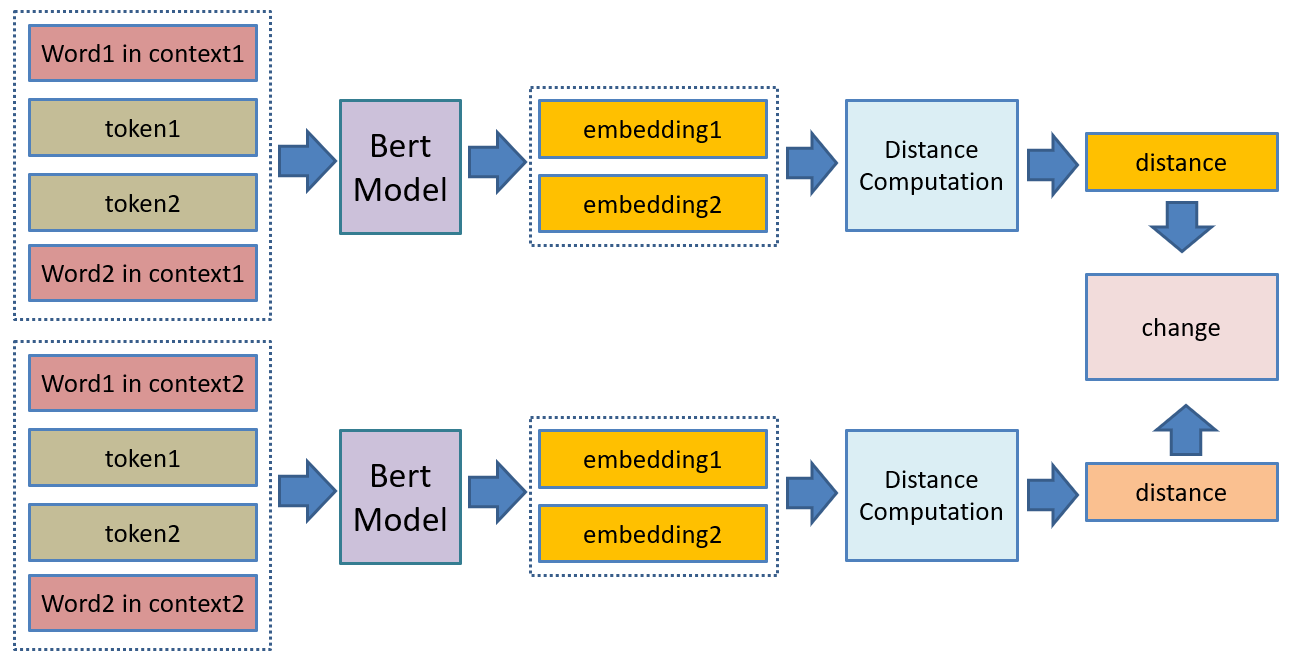}
    \captionof{figure}{Part Flow Chart of our model}
    \label{fig:model}
\end{figure}

Our model would calculate the distance by several algorithms immediately when it obtained embedding of each token, then we predict the graded effect of context in word similarity as in the following steps:
\begin{itemize}
    \item \textbf{Step1}: Choose the corresponding two embeddings of word1_context1 and word2_context1, compute the distance in several algorithms as $SC_1$. 
    \item \textbf{Step2}: Substitute the words in Step 1  with word1_context2 and word2_context2, and repeat the last step, then we get the $SC_2$.
    \item \textbf{Step3}: By subtraction, we can get the change on similarity $C = SC_1 - SC_2$
    \item \textbf{Step4}: Change the distance computing algorithm and repeat Step1 $\sim$ Step3.
    \item \textbf{Step5}: After Step1 $\sim$ Step4, we can obtain a vector of change, $C_1, C_2, \cdots, C_n$, where $n$ denotes the number of distance calculating algorithms used in our model and $w_i$ denotes the manual parameter, we get the final change \[C=\sum_{i} w_i C_i, \sum{w_i}=1\].
\end{itemize}
Here we provide a flow chart Figure \ref{fig:model} to show the process from Step1 to Step4.
\subsection{Experiment}
We trained one standard BERT Large model and one multilingual BERT Base model by MXNet \cite{chen2015mxnet}. The dataset we trained BERT Large model is \textit{openwebtext_book_corpus_wiki_en_cased}, which were maintained by GluonNLP\footnote{https://gluon-nlp.mxnet.io/model_zoo/bert/index.html}, and we trained Multilingual BERT (M-BERT)\cite{pires2019multilingual} Base model by \textit{wiki_multilingual_uncased} dataset that also provided by GluonNLP. It takes much time to train the BERT model, so we recommend utilizing the well trained BERT model from \href{https://github.com/imgarylai/bert-embedding}{bert-embedding}\footnote{We can simply use the model by pip or conda install bert-embedding}.

After configuring the models, we can follow the section \ref{overview} by giving the input from section \ref{dataset} and get the experiment results which will be introduced in Section \ref{results}. Task3 has four language tracks, namely English, Croatian, Finnish, Slovenian. We use the BERT Large model in the English track, and Multilingual BERT Base model in the other three tracks.

In section\ref{overview}, we use several algorithms to compute similarity. Here we introduce two main algorithms that used in our experiments.
\begin{itemize}
    \item Cosine Similarity that calculates the cosine of angle between two vectors.
    \[\bm{sim(w_1,w_2) = \frac{w_1\cdot w_2}{\left \|w_1\right \| \left \|w_2\right\|}=\frac{\sum_i w_{1i}w_{2i}}{\sqrt{\sum_i w_{1i}^2}\sqrt{\sum_i w_{2i}^2}}}\]
    \item Euclidean Distance that calculates the square root of square distance in each dimension.
     \[\bm{sim(w_1,w_2) = \sqrt{\sum_i(w_{1i}-w_{2i})^2}}\]
\end{itemize}
\section{Results}
\label{results}
In our experiment targeted at subtask1, the English language track uses the Bert Large model, the Euclidean distance is 0.718 and the cosine distance is 0.752, the Blend result is 0.768, and the online LB ranks second; Croatian, Finnish, and Slovenian languages all use the Multi-lingual Bert model. The Croatian language track' Euclidean distance of 0.590, the cosine distance is 0.587, the Blend result is 0.594, and the online LB ranks sixth. The Finnish language uses the Euclidean distance of 0.750, the cosine distance is 0.671, the Blend result is 0.772, and the online LB ranks 1, The Slovenian language uses a Euclidean distance of 0.576, a cosine distance of 0.603, a Blend result of 0.583, and an online LB ranking seventh. We sort the result out the following Table \ref{tab:expresult}.
\begin{table}[h]
\centering
    \begin{tabular}{cccccc}
         \toprule
         Language \& Abbr. & Model & Euclidean Dis. & Cosine & Blend & Rank \\
         \midrule
         English, En & BERT Large &  0.718 & 0.752 & 0.768 & 2 \\
         Croatian, HR & M-BERT Base &  0.590 & 0.587 & 0.594 & 6 \\
         Finnish, FI & M-BERT Base &  0.750 & 0.671 & 0.772 & 1 \\
         Slovenian, SL & M-BERT Base &  0.576 & 0.603 & 0.583 & 7 \\
         \bottomrule
    \end{tabular}
    \captionof{table}{Experiment Results}
    \label{tab:expresult}
\end{table}
\section{Conclusion}
In our paper, we propose a model that computes the similarity and similarity change by blending cosine similarity and euclidean distance, which calculated by two word embedding vectors. We firstly transform words in dataset that we introduce in section \ref{dataset}. into the word embedding vectors by BERT that we discuss in section \ref{overview}, then we calculate the distance between two vectors, finally we blend the two distances computed by different algorithms as the final predict result. In the subtask1 of task3, our team \textit{will_go} won a champion in Finnish track and the second place in English track.

\bibliographystyle{coling}
\bibliography{semeval2020}




\end{document}